\documentclass{svproc}
\usepackage{url}
\usepackage{graphicx} 
\usepackage{amssymb}
\usepackage{amsmath}
\usepackage{xcolor}
\usepackage{amstext}
\usepackage{multirow}
\usepackage{booktabs}
\usepackage{makecell}
\usepackage{comment}
\usepackage{hyperref}
\usepackage{arydshln}
\newcommand{\etal}{\textit{et al.}}

\begin{document}
\mainmatter              
\title{Zero-Shot Medical Information Retrieval via Knowledge Graph Embedding}
\author{Yuqi Wang\inst{1}\inst{3} \and Zeqiang Wang\inst{1} \and Wei Wang\inst{1} \and Qi Chen\inst{1} \and  Kaizhu Huang\inst{2} \and Anh Nguyen\inst{3} \and Suparna De\inst{4} }
\authorrunning{Wang et al.} 
%
%
\institute{
Xi'an Jiaotong-Liverpool University, Suzhou, China\\
\email{\{yuqi.wang17, zeqiang.wang22\}@student.xjtlu.edu.cn, \{wei.wang03, qi.chen02\}@xjtlu.edu.cn}
\and
Duke Kunshan University, Kunshan, China\\
\email{kaizhu.huang@dukekunshan.edu.cn}
\and
University of Liverpool, Liverpool, UK\\
\email{anh.nguyen@liverpool.ac.uk}
\and
University of Surrey, Surrey, UK\\
\email{s.de@surrey.ac.uk}
}

\maketitle             

\begin{abstract}
In the era of the Internet of Things (IoT), the retrieval of relevant medical information has become essential for efficient clinical decision-making. This paper introduces MedFusionRank, a novel approach to zero-shot medical information retrieval (MIR) that combines the strengths of pre-trained language models and statistical methods while addressing their limitations. The proposed approach leverages a pre-trained BERT-style model to extract compact yet informative keywords. These keywords are then enriched with domain knowledge by linking them to conceptual entities within a medical knowledge graph. Experimental evaluations on medical datasets demonstrate MedFusionRank's superior performance over existing methods, with promising results with a variety of evaluation metrics. MedFusionRank demonstrates efficacy in retrieving relevant information, even from short or single-term queries. 

\keywords{medical information retrieval, Internet of Things, natural language processing,  clinical decision-making, medical knowledge graph}
\end{abstract}
\section{Introduction}

The widespread adoption of the Internet of Things (IoT) has enabled the collection of large amounts of medical text data. By using IoT to identify patients, transfer information to central databases, and search for relevant medical texts such as electronic health records (EHRs) and disease-related papers, we can improve the efficiency of treatment procedures and therapeutic outcomes\cite{elhoseny2018secure,lu2021application}. For instance, the MIMIC-III\cite{johnson2016mimic} and MIMIC-IV\cite{johnson2020mimic} critical care medical databases use IoT systems to collect structured clinical data and texts. These medical texts have become the foundation for medical natural language processing, serving as corpora for pre-training large language models and embeddings\cite{alsentzer2019publicly,li2023comparative,zhang2019biowordvec}. Additionally, the use of IoT in healthcare has the potential to revolutionise patient care by providing real-time monitoring and personalised treatment plans based on individual patient data. This can lead to improved patient outcomes and reduced healthcare costs\cite{dimitrov2016medical}.

A key challenge in healthcare is enabling real-time, personalised clinical decision-making beyond traditional tasks like diagnostic classification and outcome prediction. Effective clinical decision support fundamentally relies on the ability to retrieve relevant information from massive amounts of unstructured EHR data. While earlier work in medical information retrieval relied on statistical methods like BM25\cite{robertson1995okapi} with Term Frequency-Inverse Document Frequency (TF-IDF) features, these techniques struggled with the complexity and sparsity of medical text. Medical notes exhibit pervasive synonym phenomena, with different terms like “\textit{hypertension}” and “\textit{high blood pressure}” denoting identical concepts. Abbreviations and shorthand introductions are also ubiquitous, posing difficulties for simple lexical matching.

Recently, pre-trained large language models (LLMs) like BERT\cite{devlin2018bert}, Alpaca\cite{taori2023alpaca}, and Llama\cite{touvron2023llama} have shown promise by learning generalisable representations of medical language. However, their computational overhead makes deployment directly onto resource-constrained IoT devices impractical. Training with massive LLMs requires substantial data, computing power, and memory exceeding the available on-device. Therefore, an open challenge is adapting the strengths of LLMs for medical search on embedded IoT systems. More efficient methods are needed to extract knowledge from LLMs and make it accessible for medical information retrieval on hardware-friendly architectures.

To address the aforementioned challenges, we propose a novel zero-shot information retrieval approach that integrates the strengths of statistical methods and pre-trained LLMs while mitigating their limitations. Our key insight is to leverage a pre-trained BERT-style model to extract compact yet informative keywords. These keywords are then enriched with domain knowledge by linking them to conceptual entities within a medical knowledge graph.
Our method has demonstrated promising results on two benchmark datasets, outperforming a range of existing Information Retrieval models across various evaluation metrics.

\section{Related Work}
Medical information retrieval (MIR) aims to retrieve relevant medical data from sources such as EHR. However, it faces distinct challenges that extend beyond conventional information retrieval (IR) - complex medical terminology, heterogeneous data, privacy constraints, and difficulties in system evaluation. While leveraging core IR techniques, MIR has specific requirements arising from the medical domain. In this section, we provide an overview of key IR methods that facilitate effective MIR.

\subsection{Statistical Information Retrieval}
Statistical information retrieval (Statistical IR) is a foundational approach that leverages probabilistic and statistical models to quantify the relevance of documents to user queries. This allows ranking search results by estimated relevance based on mathematical models. Popular statistical IR techniques, including vector space model\cite{christopher2008scoring}, probabilistic retrieval model\cite{sparck1972statistical}, and Okapi BM25 \cite{robertson1995okapi} rely heavily on weighted keyword matching between query and document terms. They estimate relevance using statistical signals like TF-IDF, and length normalisation. While very effective for many search tasks, these lexical similarity models have limitations. Specifically, they cannot account for semantic matching, failing to recognise synonyms and antonyms.
\subsection{Neural Information Retrieval}
Neural information retrieval (Neural IR) is a modern paradigm that leverages neural networks and deep learning techniques to overcome the limitations of statistical IR models. Neural IR models can be classified into two main types: first-stage retrieval methods and re-ranking methods.
\subsubsection{First-stage Methods}
First-stage methods aim to directly retrieve relevant documents from a large collection using neural networks. These methods can be further categorised into sparse retrieval methods and dense retrieval methods. Sparse retrieval methods use sparse word representations, such as bag-of-words or TF-IDF, as inputs to neural networks and learn to rank documents based on their similarity to queries\cite{dai2018convolutional,kim2019research}. Dense retrieval methods, on the other hand, use dense vector representations, such as word embeddings or contextual embeddings, as inputs to neural networks and learn to map queries and documents into a common semantic space where their relevance can be measured by distance metrics\cite{huang2013learning,karpukhin2020dense,shen2014learning}.
\subsubsection{Re-ranking Methods}
Re-ranking methods use neural networks to refine the initial ranking results produced by a base retriever, such as BM25 or a sparse/dense retriever. These methods can be categorised into two main approaches:
1)Re-ranking with sentence embeddings: These methods treat each document independently as an instance and learn to score its relevance to the query\cite{reimers2019sentence}. They derive vector representations for the query and each document in a separate manner, compare their embeddings and assign relevance scores. 2) Re-ranking using a cross-encoder: These methods consider each query-document pair as an instance and learn to compare their relative relevance\cite{wang2020minilm}. The cross-encoder jointly models the query and document to capture semantic matching.

\section{Methodology}
\begin{figure}
    \centering
    \includegraphics[width=\textwidth]{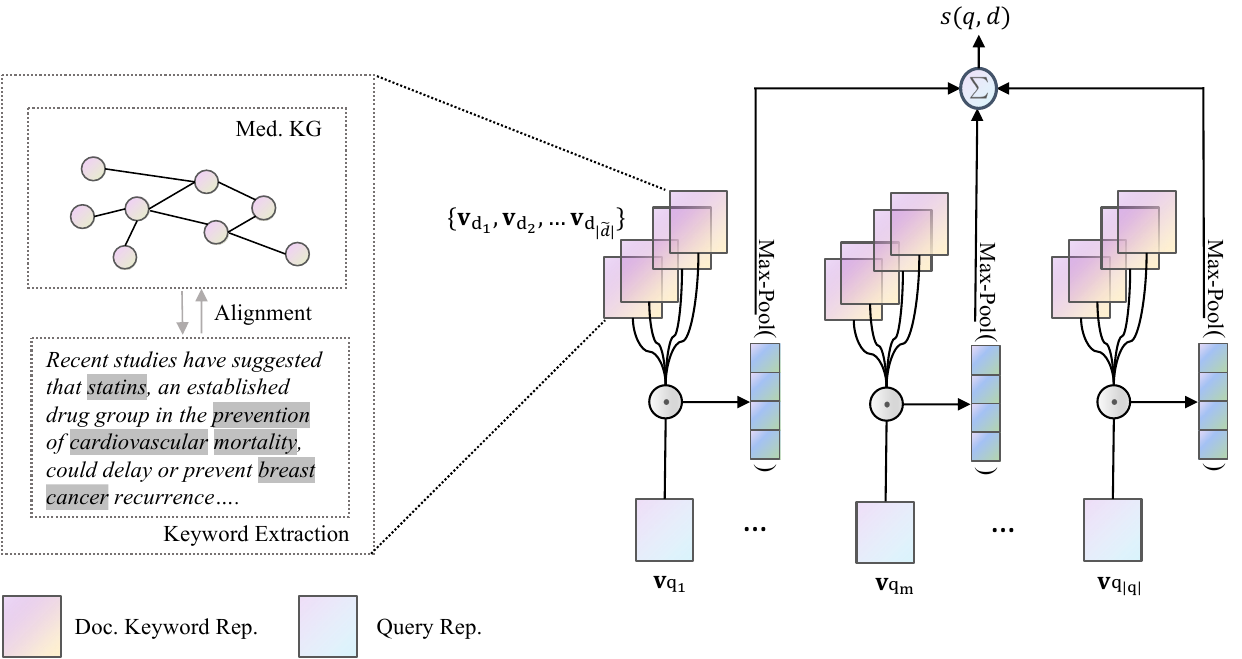}
    \caption{The overall architecture of our proposed method.}
    \label{fig:architecture}
\end{figure}
We show the overall architecture of our proposed method in Figure \ref{fig:architecture}. Specifically, it first extracts keywords from medical documents to capture semantic context. Then, medical embeddings for each keyword are constructed based on the domain-specific knowledge graph. The query and document keywords are compared in the medical embedding space and their similarity scores are aggregated to identify relevant information across query terms for retrieval. 
\subsection{Document Keyword Extraction}
Given the inherent complexity of documents within the medical domain, often encompassing multiple aspects, the necessity of pre-processing before conducting IR becomes evident. One such approach involves the extraction of keywords that aptly describe and summarise the content. By utilising a contextualised attention-based pre-trained language model, the contextual information can be effectively harnessed to discern the document's relatively significant sections. Therefore, we utilise the RoBERTa\cite{liu2019roberta} model for the initial encoding of the corpus documents. RoBERTa is a state-of-the-art language model that has demonstrated exceptional performance in various natural language processing tasks. Specifically, when dealing with a document $d$ comprised of $k$ words, denoted as $d = \{d_1, ... d_k\}$, we leverage the RoBERTa encoding function, $f(\cdot;\theta)$, to transform all the words into a coherent and meaningful semantic space, i.e. 
\begin{equation}
\{\mathbf{h}_{<s>},\mathbf{h}_{d_1}, ...\mathbf{h}_{d_k},\mathbf{h}_{</s>}\} = f(\{<s>,d_1, ... d_k, </s>\};\theta)
\end{equation}
where $\mathbf{h}_{d_i}$ is the representation of the $i$-th word in RoBERTa embedding space. $<s>$ and $</s>$ are two special tokens indicating the start and the end positions in the document, respectively.
This process enables us to capture the intricate contextual relationships and nuances present within the document. 

The comprehensive essence of the document is commonly encapsulated within the hidden state of the special token $<s>$; in order to estimate the significance of individual words within the document, we compute the cosine similarity between the representation of the special token $<s>$ and the representation of each word. We take the top $K$ ranking words based on their similarity scores, and extract those as the key keywords for the document $d$. This process is articulated as follows:
\begin{equation}
    \tilde{d} = \underset{d_i \in d} {\operatorname{top}K}\left [\operatorname{Sim}\left(\textbf{h}_{d_i}, \textbf{h}_{<s>}\right)\right]
\end{equation}
where $\tilde{d}$ is the keyword set for document $d$, $\operatorname{Sim}(\cdot)$ is the cosine similarity function. Based on our observation, the top 20 keywords can effectively capture the core semantic content of a document. Hence, we set the number of extracted keywords ($K$) to 20.
\subsection{Medical Embedding Construction}
In our work, the challenge posed by zero-shot IR is significant, primarily due to the absence of any prior exposure of the model to the medical domain. In this case, a crucial approach involves enhancing each keyword in the keyword set $\tilde{d}$ with relevant background information. This enrichment encompasses additional context, definitions, and pertinent details sourced from the medical field. In this endeavour, the Medical Subject Headings (MeSH)\cite{lipscomb2000medical} knowledge graph emerges as an exceptional resource. MeSH is a meticulously structured and high-quality knowledge graph that encompasses a vast spectrum of medical concepts along with their relationships. For instance,  the relation``\textit{treatment}'' connects the two concepts ``\textit{cancer}'' and ``\textit{chemotherapy}''. This indicates that chemotherapy is a type of treatment commonly used for cancer patients.

To harness the knowledge from MeSH, a method called Node2Vec\cite{grover2016node2vec} can be used to generate medical embeddings. The main idea is to treat this graph as a network, where nodes are concepts and edges represent relationships between concepts \cite{zhang2019biowordvec}. This method utilises random walks and learns latent representations of nodes that maximise the probability of the sampled walks. The objective function $J$ for constructing the medical embeddings can be written as follows:
\begin{equation}
    J=\max \left[\frac{1}{T} \sum_{i = 1}^T \sum_{v_j \in \mathcal{C}(v_i)} \log p\left(v_j \mid v_i\right)\right]
\end{equation}
where $T$ is the number of the MeSH concepts and $\mathcal{C}(v_i)$ is a set containing surrounding words of $v_i$ based on random walks in the knowledge graph. For this study,  alignment between the keyword set $\tilde{d}$, the query $q$, and concepts in the MeSH knowledge graph were performed by matching keywords with concept names. This simple lexical approach to entity linking was chosen for its simplicity. However, it has known limitations, such as ambiguity and lack of semantic matching. Future work should explore more sophisticated techniques to deal with the issue.
\subsection{Retrieval with Medical Knowledge}
By acquiring all the medical embeddings for document keywords from a corpus in the MeSH knowledge graph embedding space through an offline process, we can retrieve relevant information for each word from a given human-generated query in an efficient manner. In particular, each query term can focus on each word in the document to identify the most relevant information in the document that can be retrieved by that specific query word. We aggregate all the relevance scores for each query term during the retrieval process, i.e.
\begin{equation}
    s(q,d) = \sum_{i=1}^{|q|} \max
    _{j=1}^{|\tilde{d}|} \left[\mathbf{v}_{q_i} \odot\mathbf{v}_{d_j} \right]
\end{equation}
where $|q|$ and $|\tilde{d}|$ are the number of words in the query and document keyword set, respectively. $\odot$ is the dot product operation symbol. $\mathbf{v}_{q_i}$ and $\mathbf{v}_{d_j}$ are corresponding medical embeddings for the $i$-th word in the query and $j$-th word in the document keyword set.

One clear limitation of Retrieval with Medical Knowledge is the equal weighting given to documents whose keyword sets contain query terms, regardless of term frequency. Despite the inclusion of background knowledge corresponding to each word in the document’s keywords, factors such as term frequency should also be considered. BM25 \cite{robertson1995okapi} is a commonly used unsupervised ranking function, incorporating lexical aspects and statistical information to improve scoring. Leveraging medical embeddings enables the retrieval of candidate-relevant documents while applying BM25, which can further refine the ranking of those initial results by incorporating term frequency statistics. Therefore, we propose fusing the scores yielded by both approaches to improve overall performance, i.e.
\begin{equation}
    \hat{s}(q,d) = \left\{\begin{array}{lll}
			s(q,d) + s^{\prime}(q,d) &&\exists s^{\prime}(q,d)\\ \\
			s(q,d)  &&\nexists s^{\prime}(q,d)
		\end{array}\right.
\end{equation}
where $s^{\prime}(q,d)$ represents the BM25 score assigned to a given query $q$ and document $d$. $\hat{s}(q,d)$ is the final score after the fusion.
\section{Results and Evaluation}
We evaluated the performance of our proposed models on two medical datasets: NFCorpus \cite{boteva2016full} and SCIFACT \cite{wadden2020fact}. Both focus on retrieving medical abstracts relevant to search queries. The abstracts are written in technical medical terminology, mostly from PubMed. For each dataset, a range of metrics, including Mean Reciprocal Rank (MRR), Precision, normalised Discounted Cumulative Gain (nDCG), Precision (P) and Recall (R), was employed for a thorough evaluation. Our model was compared against several first-stage retrievers and BM25-based re-rankers to assess its effectiveness.
\subsection{Baseline Models}
\subsubsection{First-stage Retrievers}
\begin{itemize}
    \item \textbf{BioLinkBERT} \cite{raj2022biosimcse} and \textbf{S-BERT} \cite{reimers2019sentence}: These are two BERT-based models that generate sentence embeddings using siamese networks. While S-BERT was pre-trained on a general domain question-answering dataset to create universal semantic embeddings, BioLinkBERT utilises contrastive learning on medical texts from PubMed to produce embeddings specialised for the medical domain.
    \item \textbf{DocT5Query} \cite{nogueira2019document}: It leverages a pre-trained T5\cite{raffel2020exploring} model to generate synthetic queries based on the document for text enrichment before indexing. 
    \item \textbf{DeepCT} \cite{deepct}: It employs the BERT model to estimate the weight of each word in the context of the document. These BERT-derived weights are then used to modify the term frequencies of the words. 
    \item \textbf{BM25} \cite{robertson1995okapi}: It is a traditional unsupervised ranking function. The basic idea is that a more relevant document will contain more of the query terms, and multiple occurrences of a term can indicate higher relevance.
\end{itemize}
\subsubsection{BM25-based re-rankers}
\begin{itemize}
    \item \textbf{S-BERT} \cite{reimers2019sentence}: We used the same S-BERT model as described previously to re-rank the top 100 candidate documents retrieved in the first-stage for each query.
    \item \textbf{Cross Encoder} \cite{wang2020minilm}: It passes both the query and document sentence simultaneously to a Transformer network, producing an output value between 0 and 1, which indicates the relevance of the sentence pair. In reference to a study by Thakur \etal  \cite{thakur2021beir},  it is highlighted that MiniLM demonstrates the best performance. Therefore, we evaluate the performance when using MiniLM as the Cross Encoder for re-ranking.

\end{itemize}
\subsection{Main Results}
\begin{table}[ht!]
    \centering
    \resizebox{\linewidth}{!}{
        \begin{tabular}{lcccccccc}
        \toprule
         \multirow{2}{*}[-0.8ex]{\textbf{Method}} & \multicolumn{4}{c}{\textbf{NFCorpus}} & \multicolumn{4}{c}{\textbf{SCIFACT}}\\
         \cmidrule(r){2-5} \cmidrule(r){6-9}
         & \textbf{MRR} & \textbf{P@10} & \textbf{nDCG@10} & \textbf{R@1k} & \textbf{MRR} & \textbf{P@10} & \textbf{nDCG@10} & \textbf{R@1k}\\
        \midrule
        \multicolumn{9}{c}{First-stage Retrievers}\\
        \hdashline
        BioLinkBERT & 0.329 & 0.132 & 0.173 & 0.532 & 0.519 & 0.076 & 0.550 & 0.979\\
        S-BERT & 0.501 & 0.218 & 0.300 & 0.574 & 0.570 & 0.082 & 0.596 & 0.959\\
        DocT5Query\dag & - & - & 0.328 & -& - & - & 0.675 & -\\
        DeepCT\dag & - & - & 0.283 & - & - & - & 0.630 & -\\
        BM25 & 0.537 & 0.233 & 0.325 & 0.372 & 0.635 & 0.088 & 0.665 & 0.980\\
        \midrule
        \multicolumn{9}{c}{BM25-based Re-rankers}\\
        \hdashline
        Cross Encoder &\textbf{0.591} & 0.244 & 0.350 & 0.250 & 0.662 & 0.091 & 0.688 & 0.908\\
        S-BERT & 0.430 & 0.170 & 0.232 & 0.229 & 0.539 & 0.081 & 0.568 & 0.864\\  
        \midrule
        \multicolumn{9}{c}{Our Proposed Models}\\
        \hdashline
        MedRetriever \textasteriskcentered& 0.499 & 0.222 & 0.298 & \textbf{0.644} & 0.540 & 0.083 & 0.581 & \textbf{0.990}\\
        MedFusionRank & 0.552 & \textbf{0.262} & \textbf{0.357} &\textbf{0.644} & \textbf{0.673} & \textbf{0.094}& \textbf{0.705} & \textbf{0.990}\\
        \bottomrule
        \end{tabular}
    }
    \caption{Performances of first-stage retrievers, BM25-based re-rankers and our proposed models. \dag The results were cited from \cite{thakur2021beir}. \textasteriskcentered MedRetriever refers to our proposed method as a standalone approach, distinct from its fusion with BM25.}
    \label{tab:main_results}
\end{table}
The main retrieval results are illustrated in Table \ref{tab:main_results}. It demonstrates that BM25 is an effective baseline for zero-shot IR compared with bi-encoders such as S-BERT and BioLinkBERT. BM25 ranking alone achieves reasonable performance, which can be further improved by re-ranking using a cross-encoder model. This two-stage ranking pipeline achieves the best MRR results on the NFCorpus dataset. However,  re-ranking based on BM25 has limitations stemming from BM25’s dependence on exact term matching, which can cause relevant documents to be excluded from consideration during later re-ranking stages.

A noteworthy scenario emerged where the precision of MedRetriever at the top 1000 exhibited favourable results among all the baseline retrievers. In contrast, the nDCG at the top 10 demonstrated comparatively suboptimal performance. This disparity between precision and nDCG metrics suggests that although the MedRetriever is capable of retrieving a fair proportion of relevant documents overall, it struggles to rank the most relevant documents at the very top of the list. When we combine scores from two methods, MedRetriever and BM25, the results consistently outperformed nearly all of the baseline methods across all evaluation metrics.
\subsection{Out-of-Vocabulary strategy}
\begin{table}[ht!]
    \centering
    \resizebox{\linewidth}{!}{
        \begin{tabular}{lcccccccc}
        \toprule
         \multirow{2}{*}[-0.8ex]{\textbf{Method}} & \multicolumn{4}{c}{\textbf{NFCorpus}} & \multicolumn{4}{c}{\textbf{SCIFACT}}\\
         \cmidrule(r){2-5} \cmidrule(r){6-9}
         & \textbf{MRR} & \textbf{P@10} & \textbf{nDCG@10} & \textbf{R@1k} & \textbf{MRR} & \textbf{P@10} & \textbf{nDCG@10} & \textbf{R@1k}\\
        \midrule

        Prefix Approx. & 0.552 & 0.262 & 0.357 & \textbf{0.644} & 0.673 & 0.094& 0.705 & 0.990\\
        CharLSTM & \textbf{0.553} & \textbf{0.263} & \textbf{0.358} & 0.643 & \textbf{0.684} & 0.094 & \textbf{0.713} & 0.990\\
        \bottomrule
        \end{tabular}
    }
    \caption{Performances of using different out-of-vocabulary strategies for MedFusionRank}
    \label{tab:oov}
\end{table}
To handle out-of-vocabulary (OOV) words, this work incorporates two strategies: Prefix Approximation and a Character-level Long Short-Term Memory network (CharLSTM). Prefix Approximation, originally proposed in \cite{speer2017conceptnet}, identifies the longest common prefix between an OOV word and in-vocabulary words, then averages all embeddings sharing that prefix to represent the OOV term. On the other hand, the CharLSTM learns sequential character-level features of in-vocabulary words to construct a non-linear mapping from character sequences to medical embeddings. As depicted in Table \ref{tab:oov}, the CharLSTM achieves better overall performance compared to Prefix Approximation. This indicates that modelling the sequential patterns and characters of medical terminology plays a more vital role in estimating representations for OOV words in this domain.

\subsection{Case Study}

\begin{table}[ht]
\begin{tabular}{c|ccccc}
\toprule
Query                                                      & \multicolumn{5}{c}{Keywords in retrieved document}                                                                     \\ \midrule
\multirow{4}{*}{\textit{zoloft}}                          & \textit{depression}   & \textit{depressive}     & \textit{antidepressants} & \textit{exercise}   & \textit{sertraline} \\
                                                           & \textit{aerobic}      & \textit{therapy}        & \textit{anxiety}         & \textit{treatment}  & \textit{medication} \\
                                                           & \textit{therapeutic}  & \textit{50}             & \textit{disorders}       & \textit{mental}     & \textit{older}      \\
                                                           & \textit{effects}      & \textit{67}             & \textit{rating}          & \textit{mdd}        & \textit{diagnostic} \\ \midrule
\multicolumn{1}{l|}{\multirow{4}{*}{\textit{myelopathy}}} & \textit{spinal}       & \textit{sclerotic}      & \textit{paraplegia}      & \textit{cobalamin}  & \textit{spine}      \\
\multicolumn{1}{l|}{}                                     & \textit{vegetarian}   & \textit{vegan}          & \textit{subacute}        & \textit{cervical}   & \textit{vitamin}    \\
\multicolumn{1}{l|}{}                                     & \textit{degeneration} & \textit{hypertonia}     & \textit{diagnosed}       & \textit{reflexia}   & \textit{impairment} \\
\multicolumn{1}{l|}{}                                     & \textit{paresthesias} & \textit{rehabilitative} & \textit{hypotrophy}      & \textit{neurogenic} & \textit{diet}       \\ \bottomrule
\end{tabular}
\caption{Keywords in the retrieved document based on a single term as query}\label{tab:casestudy}
\end{table}
To further evaluate the performance of our proposed model, we conducted a case study using short, single-term queries common in human searches. Statistical matching models like BM25 often struggle with these sparse queries, as the single terms may not exist in the corpus. As shown in Table \ref{tab:casestudy}, the sample query terms “\textit{zoloft}” and “\textit{myelopathy}” did not appear in any documents. However, our proposed model successfully retrieved relevant documents with medical concepts from the knowledge graph, ranking pertinent documents in the top 10 results for both queries. 

In the first example, “\textit{zoloft}” is an antidepressant medication. Therefore, “\textit{depression}”, “\textit{depressive}”, and “\textit{anxiety}” are closely connected to “\textit{zoloft}” since the medication aims to alleviate the symptoms associated with these conditions. In another example, “\textit{myelopathy}” is a spinal cord pathology that can result from vitamin deficiency, spinal degeneration, or cord compression. The keywords “\textit{spinal}”, “\textit{spine}”, “\textit{vitamin}” and “\textit{degeneration}” from the retrieved document could be relevant to the query.

This case study highlights the potential of our proposed model to improve the search relevancy of short user queries. Our model effectively utilised associated medical concepts to match user information needs.

\section{Conclusion and Future Work}
In this paper, we have presented MedFusionRank, a novel zero-shot MIR approach that integrates the strengths of statistical methods and pre-trained language models. Our key insight is to leverage a pre-trained BERT-style model to extract compact yet informative keywords. These keywords are then enriched with domain knowledge by linking them to conceptual entities within a medical knowledge graph.

Our experiments on two benchmark medical datasets demonstrate that MedFusionRank achieves promising results, outperforming a range of existing models across various evaluation metrics. The case study also reveals MedFusionRank's ability to retrieve relevant documents even for short or single-term queries.

There are several exciting directions for future work. First, we plan to expand the coverage of our medical knowledge graph using more comprehensive knowledge resources. Second, we intend to explore more sophisticated entity-linking techniques beyond simple lexical matching. Third, to enable deployment on resource-constrained IoT devices, we will construct a vector database of the encoded document embeddings and load it directly onto the target hardware. This will circumvent the need for inference-time encoding and drastically reduce retrieval latency and memory overhead. Finally, we aim to implement an end-to-end prototype for real-time clinical decision support on medical IoT devices.

\section*{Acknowledgement}
 We would like to acknowledge the financial support provided by the Postgraduate Research Scholarship (PGRS) at Xi’an Jiaotong-Liverpool University (contract number PGRS2006013). Additionally, this research has received partial funding from the Jiangsu Science and Technology Programme (contract number BK20221260).

%
%
\bibliography{author}
\bibliographystyle{spmpsci}

\end{document}